\documentstyle[anlp,fancyheadings]{article}

\title{Adaptive Sentence Boundary Disambiguation}

\author{David D. Palmer \\
CS Division, 387 Soda Hall \#1776\\
University of California, Berkeley\\
Berkeley, CA 94720-1776\\
{\it dpalmer@cs.berkeley.edu}\\
\And Marti A. Hearst \\
Xerox PARC \\
3333 Coyote Hill Rd\\
Palo Alto, CA 94304\\
{\it hearst@parc.xerox.com}\\}

\begin{document}
\maketitle
\bibliographystyle{acl}

\begin{abstract}
Labeling of sentence boundaries is a necessary prerequisite for many
natural language processing tasks, including part-of-speech tagging
and sentence alignment.  End-of-sentence punctuation marks are
ambiguous; to disambiguate them most systems use brittle,
special-purpose regular expression grammars and exception rules.  As
an alternative, we have developed an efficient, trainable algorithm
that uses a lexicon with part-of-speech probabilities and a
feed-forward neural network.  This work demonstrates the feasibility
of using prior probabilities of part-of-speech assignments, as opposed
to words or definite part-of-speech assignments, as contextual
information.   After training for less than one minute, the method
correctly labels over 98.5\% of sentence boundaries in a corpus of
over 27,000 sentence-boundary marks.  We show the method to be
efficient and easily adaptable to different text genres, including
single-case texts.
\end{abstract}

\thispagestyle{fancy}
\headrulewidth 	      0.0pt
\footrulewidth	      0.0pt
\chead{\tt \large
In the Proceedings of the ANLP '94, Stuttgart, Germany, October 1994.}
\pagestyle{plain}

\section{Introduction}
\label{intro}

Labeling of sentence boundaries is a necessary prerequisite for many
natural language processing (NLP) tasks, including part-of-speech tagging
\cite{church88},~\cite{cutting91}, and sentence alignment
\cite{gale93},~ \cite{kay93}.  End-of-sentence punctuation marks are
ambiguous; for example, a period can denote an abbreviation,
the end of a sentence, or both, as shown in the examples below:

\begin{itemize}
\item[(1)] {\it The group included Dr. J.M. Freeman and T. Boone Pickens Jr.}
\item[(2)] {\it ``This issue crosses party lines and crosses philosophical lines!'' said Rep. John Rowland (R., Conn.).}
\end{itemize}

Riley \shortcite{riley89} determined that in the Tagged Brown corpus
\cite{francis82} about 90\% of periods occur at the end of sentences,
10\% at the end of abbreviations, and about 0.5\% as both
abbreviations and sentence delimiters.  Note from example (2) that
exclamation points and question marks are also ambiguous, since they
too can appear at locations other than sentence boundaries.

Most robust NLP systems, e.g., Cutting et al. \shortcite{cutting91},
find sentence delimiters by tokenizing the text stream and applying a
regular expression grammar with some amount of look-ahead, an
abbreviation list, and perhaps a list of exception rules.  These
approaches are usually hand-tailored to the particular text and rely
on brittle cues such as capitalization and the number of spaces
following a sentence delimiter.  Typically these approaches use only
the tokens immediately preceding and following the punctuation mark to
be disambiguated.  However, more context can be necessary, such as
when an abbreviation appears at the end of a sentence, as seen in
(3a-b):

\begin{itemize}
\item[(3a)] {\it It was due Friday by 5 p.m.  Saturday would be too late.}
\item[(3b)] {\it She has an appointment at 5 p.m. Saturday to get her car
fixed.} 
\end{itemize}

\noindent
or when punctuation occurs in a subsentence within quotation marks or
parentheses, as seen in Example (2).

Some systems have achieved accurate boundary determination by applying
very large manual effort.  For example, at Mead Data Central, Mark
Wasson and colleagues, over a period of 9 staff months, developed a
system that recognizes special tokens (e.g., non-dictionary terms such
as proper names, legal statute citations, etc.)~as well as sentence
boundaries.  From this, Wasson built a stand-alone boundary recognizer
in the form of a grammar converted into finite automata with 1419
states and 18002 transitions (excluding the lexicon).  The resulting
system, when tested on 20 megabytes of news and case law text, achieved
an accuracy of 99.7\% at speeds of 80,000 characters per CPU second on
a mainframe computer.  When tested against upper-case legal text
the algorithm still performed very well, achieving accuracies of
99.71\% and 98.24\% on test data of 5305 and 9396 periods,
respectively.  It is not likely, however, that the results would be
this strong on lower-case data.\footnote{All information about
Mead's system is courtesy of a personal communication with Mark
Wasson.}

Humphrey and Zhou \shortcite{humphrey89} report using a feed-forward neural
network to disambiguate periods, although they use a regular grammar
to tokenize the text before training the neural nets, and achieve an
accuracy averaging 93\%.\footnote{Accuracy results were obtained
courtesy of a personal communication with Joe Zhou.}

Riley \shortcite{riley89} describes an approach that uses regression trees
\cite{breiman84} to classify sentence boundaries according to the
following features:

\begin{itemize}
\item[] Probability[word preceding ``.'' occurs at end of sentence]
\item[] Probability[word following ``.'' occurs at beginning of sentence]
\item[] Length of word preceeding ``.''
\item[] Length of word after ``.''
\item[] Case of word preceeding ``.'': Upper, Lower, Cap, Numbers
\item[] Case of word following ``.'': Upper, Lower Cap, Numbers
\item[] Punctuation after ``.'' (if any)
\item[] Abbreviation class of words with ``.''
\end{itemize}

The method uses information about one word of context on either side
of the punctuation mark and thus must record, for every word in the
lexicon, the probability that it occurs next to a sentence boundary.
Probabilities were compiled from 25 million words of pre-labeled
training data from a corpus of AP newswire.  The results were tested on the
Brown corpus achieving an accuracy of 99.8\%.\footnote{Time for
training was not reported, nor was the amount of the Brown corpus
against which testing was performed; we assume the entire Brown corpus
was used.}

M\"{u}ller \shortcite{mueller80} provides an exhaustive analysis of
sentence boundary disambiguation as it relates to lexical endings and
the identification of words surrounding a punctuation mark, focusing
on text written in English.  This approach makes multiple passes
through the data and uses large word lists to determine the positions
of full stops.  Accuracy rates of 95-98\% are reported for this method
tested on over 75,000 scientific abstracts.  (In contrast to Riley's
Brown corpus statistics, M\"{u}ller reports sentence-ending to
abbreviation ratios ranging from 92.8\%/7.2\% to 54.7\%/45.3\%. This
implies a need for an approach that can adapt flexibly to the
characteristics of  different
text collections.)

Each of these approaches has disadvantages to overcome.  We propose
that a sentence-boundary disambiguation algorithm have the following
characteristics:

\begin{itemize}

\item The approach should be robust, and should not require a hand-built
grammar or specialized rules that depend on capitalization, multiple
spaces between sentences, etc.  Thus, the approach should adapt easily
to new text genres and new languages.

\item The approach should train quickly on a small training set
and should not require excessive storage overhead.

\item The approach should be very accurate and 
efficient enough that it does not noticeably slow down 
text preprocessing.

\item The approach should be able to specify ``no opinion'' on
cases that are too difficult to disambiguate, rather than making 
underinformed guesses.

\end{itemize}

In the following sections we present an approach that meets each of
these criteria, achieving performance close to solutions that
require manually designed rules, and behaving more
robustly.  Section \ref{algorithm} describes the algorithm, Section
\ref{results} describes some experiments that evaluate the
algorithm, and Section \ref{summary} summarizes the paper and
describes future directions.

\section{Our Solution}
\label{algorithm}

We have developed an efficient and accurate automatic sentence
boundary labeling algorithm which overcomes the limitations of
previous solutions.  The method is easily trainable and adapts to new
text types without requiring rewriting of recognition rules.  The core
of the algorithm can be stated concisely as follows: the
part-of-speech probabilities of the tokens surrounding a punctuation
mark are used as input to a feed-forward neural network, and the
network's output activation value determines what label to assign to
the punctuation mark.

The straightforward approach to using contextual information is to
record for each word the likelihood that it appears before or after a
sentence boundary.  However, it is expensive to obtain probabilities
for likelihood of occurrence of all individual tokens in the positions
surrounding the punctuation mark, and most likely such information 
would not be useful to any subsequent processing steps in an NLP
system.  Instead, we use probabilities for the part-of-speech
categories of the surrounding tokens, thus making training faster and
storage costs negligible for a system that must in any case record
these probabilities for use in its part-of-speech tagger.

This approach appears to incur a cycle: because most
part-of-speech taggers require pre-determined sentence boundaries,
sentence labeling must be done before tagging.  But if sentence
labeling is done before tagging, no part-of-speech assignments are
available for the boundary-determination algorithm.  Instead of
assigning a single part-of-speech to each word, our algorithm uses
{\it the prior probabilities} of all parts-of-speech for that word.
This is in contrast to Riley's method \cite{riley89} which requires
probabilities to be found for every lexical item (since it records
the number of times every token has been seen before and after a
period).  Instead, we suggest making use of the unchanging prior
probabilities for each word already stored in the system's lexicon.

The rest of this section describes the algorithm in more detail.

\subsection{Assignment of Descriptors}

The first stage of the process is lexical analysis, which breaks the
input text (a stream of characters) into tokens.  Our implementation
uses a slightly-modified version of the tokenizer from the PARTS
part-of-speech tagger \cite{church88} for this task.  A token can be a
sequence of alphabetic characters, a sequence of digits (numbers
containing periods acting as decimal points are considered a single
token), or a single non-alphanumeric character.  A lookup module then
uses a lexicon with part-of-speech tags for each token.  This lexicon
includes information about the frequency with which each word occurs
as each possible part-of-speech.  The lexicon and the frequency counts
were also taken from the PARTS tagger, which derived the counts from
the Brown corpus
\cite{francis82}. For the word {\sl adult}, for example, the lookup
module would return the tags ``JJ/2 NN/24,'' signifying that the word
occurred 26 times in the Brown corpus -- twice as an adjective and 24
times as a singular noun.

The lexicon contains 77  part-of-speech tags, which we map
into 18 more general categories (see Figure 1).  For example, the
tags for present tense verb, past participle, and modal verb 
all map into the more general ``verb'' category.  For a given
word and category, the frequency of the category is the sum of the
frequencies of all the tags that are mapped to the category for that
word.  The 18 category frequencies for the word are then converted to
probabilities by dividing the frequencies for each category by the
total number of occurrences of the word.

For each token that appears in the input stream, a descriptor array is
created consisting of the 18 probabilities as well as two additional
flags that indicate if the word begins with a capital letter and if it
follows a punctuation mark.

\begin{figure*}
\centering
\begin{tabular} {l l l l} 
noun & verb & article & modifier  \\
conjunction & pronoun & preposition & proper noun \\
number & comma or semicolon & left parentheses & right parentheses \\
non-punctuation character & possessive & colon or dash & abbreviation \\
sentence-ending punctuation & others & & \\
\end{tabular}
\caption{Elements of the Descriptor Array assigned to each incoming
token. }
\end{figure*}

\subsection{The Role of the Neural Network}

We accomplish the disambiguation of punctuation marks using a
feed-forward neural network trained with the back propagation
algorithm \cite{hertz91}.  The network accepts as input $k * 20$ input
units, where $k$ is the number of words of context surrounding an
instance of an end-of-sentence punctuation mark (referred to in this
paper as ``k-context''), and $20$ is the number of elements in the
descriptor array described in the previous subsection.  The input
layer is fully connected to a hidden layer consisting of $j$ hidden
units with a sigmoidal squashing activation function.  The hidden
units in turn feed into one output unit which indicates the results of
the function.\footnote{The context of a punctuation mark can be
thought of as the sequence of tokens preceding and following it.  Thus
this network can be thought of roughly as a Time-Delay Neural Network
(TDNN) \cite{hertz91}, since it accepts a sequence of inputs and is
sensitive to positional information within the sequence.  However,
since the input information is not really shifted with each time step,
but rather only presented to the neural net when a punctuation mark is
in the center of the input stream, this is not technically a TDNN.}

The output of the network, a single value between 0 and 1, represents
the strength of the evidence that a punctuation mark occurring in its
context is indeed the end of the sentence.  We define two adjustable
sensitivity thresholds $t_0$ and $t_1$, which are used to classify the
results of the disambiguation.  If the output is less than $t_0$, the
punctuation mark is not a sentence boundary; if the output is greater
than or equal to $t_1$, it is a sentence boundary.  Outputs which fall
between the thresholds cannot be disambiguated by the network and are
marked accordingly, so they can be treated specially in later
processing.  When $t_0 = t_1$, every punctuation mark is labeled as
either a boundary or a non-boundary.

To disambiguate a punctuation mark in a k-context, 
a window of $k+1$ tokens and their
descriptor arrays is maintained as the input text is read.  The first
$k/2$ and final $k/2$ tokens of this sequence represent the context in
which the middle token appears.  If the middle token is a potential
end-of-sentence punctuation mark, the descriptor arrays for the
context tokens are input to the network and the output result
indicates the appropriate label, subject to the thresholds $t_0$ and
$t_1$. 

Section \ref{results} 
describes experiments which vary the size of $k$
and the number of hidden units.

\subsection{Heuristics}

A connectionist network can discover patterns in the input data
without using explicit rules, but the input must be structured to
allow the net to recognize these patterns.  Important factors in the
effectiveness of these arrays include the mapping of part-of-speech
tags into categories, and assignment of parts-of-speech to words not
explicitly contained in the lexicon.

As previously described, we map the part-of-speech tags in the lexicon
to more general categories.  This mapping is, to an extent,
dependent on the range of tags and on the language being analyzed.
In our experiments, when all verb forms
in English are placed in a single category, the results are strong
(although we did not try alternative mappings).  We speculate,
however, that for languages like German, the verb forms will
need to be separated from each other, as certain forms occur
much more frequently at the end of a sentence than others do.  
Similar issuse may arise in other languages.

Another important consideration is classification of words not present
in the lexicon, since most texts contain infrequent words.
Particularly important is the ability to recognize tokens that are
likely to be abbreviations or proper nouns. M\"{u}ller \shortcite{mueller80}
gives an argument for the futility of trying to compile an exhaustive list
of abbreviations in a language, thus implying the need to recognize
unfamiliar abbreviations.  We implement several techniques to accomplish
this.  For example, we attempt to identify
initials by assigning an ``abbreviation'' tag to all sequences of
letters containing internal periods and no spaces.  This finds
abbreviations like ``J.R.'' and ``Ph.D.''  Note that the final period
is a punctuation mark which needs to be disambiguated, and is
therefore not considered part of the word.

A capitalized word is not necessarily a proper noun, even when it
appears somewhere other than in a sentence's initial position (e.g.,
the word ``American'' is often used as an adjective).  We require a
way to assign probabilities to capitalized words that appear in the
lexicon but are not registered as proper nouns.  We use a simple
heuristic: we split the word's probabilities, assigning a $0.5$
probability that the word is a proper noun, and dividing the remaining
$0.5$ according to the proportions of the probabilities of the parts
of speech indicated in the lexicon for that word.

Capitalized words that do not appear in the lexicon at all are generally
very likely to be proper nouns; therefore, they are assigned a proper noun
probability of $0.9$, with the remaining $0.1$ probability distributed
equally among all the other parts-of-speech.  These simple assignment
rules are effective for English, but would need to be slightly
modified for other languages with different capitalization rules
(e.g., in German all nouns are capitalized).

\section{Experiments and Results}
\label{results}

We tested the boundary labeler on a large body of text 
containing 27,294 potential sentence-ending punctuation marks
taken from the Wall Street Journal portion of the ACL/DCI collection
\cite{church91c}.   No preprocessing was performed on the test
text, aside from removing unnecessary headers and correcting existing
errors.  (The sentence boundaries in the WSJ text had been previously
labeled using a method similar to that used in PARTS and is described
in more detail in \cite{liberman92}; we found and corrected several
hundred errors.)  We trained the weights in the neural network with a 
back-propagation algorithm on a training set of 573 items from the same
corpus.  To increase generalization of training, a separate 
cross-validation set (containing 258 items also from the same corpus) 
was also fed through the network, but the weights were not trained on
this set.  When the cumulative error of the items in the cross-validation
set reached a minimum, training was stopped.  Training was done in
batch mode with a learning rate of $0.08$.
The entire training procedure required less than one minute on a 
Hewlett Packard 9000/750 Workstation.  This should be contrasted with Riley's
algorithm which required 25 million words of training data in order to
compile probabilities.

If we use Riley's statistics presented in Section \ref{intro}, we can
determine a lower bound for a sentence boundary disambiguation
algorithm: an algorithm that always labels a period as a sentence
boundary would be correct 90\% of the time; therefore, any method must
perform better than 90\%.  In our experiments, performance was very
strong: with both sensitivity thresholds set to $0.5$, the network
method was successful in disambiguating $98.5\%$ of the punctuation
marks, mislabeling only 409 of 27,294.  These errors fall into two
major categories: (i)``false positive'': the method erroneously
labeled a punctuation mark as a sentence boundary, and (ii) ``false
negative'': the method did not label a sentence boundary as
such.  See Table \ref{errors} for details.

\begin{table}[h]
\centering
\begin{tabular}{| l |} \hline
224 (54.8\%) false positives \\
185 (45.2\%) false negatives  \\ \hline 
409 total errors out of 27,294 items \\ \hline
\end{tabular}
\caption{\protect{Results of testing on 27,294 mixed-case
items;  $t_0 = t_1 = 0.5$, 6-context, 2 hidden units.}
\label{errors}}
\end{table}

The 409 errors from this testing run can be decomposed into the
following groups:

\begin{center}
\begin{itemize}
\item[]\begin{itemize}
\parsep         -3pt
\itemsep        -3pt
\item[37.6\%] false positive at an abbreviation within a title or name, usually
	because the word following the period exists in the lexicon
	with other parts-of-speech ({\sl Mr. Gray, Col. North,
	Mr. Major, Dr. Carpenter, Mr. Sharp}).  Also included
	in this group are items such as {\sl U.S. Supreme Court}
	or {\sl U.S. Army}, which are sometimes mislabeled
	because {\sl U.S.} occurs very frequently 
	at the end of a sentence as well.

\item[22.5\%] false negative due to an abbreviation at the end of a sentence,
	most frequently {\sl Inc., Co., Corp.}, or {\sl U.S.}, which all occur
	within sentences as well.

\item[11.0\%] false positive or negative due to a sequence of characters
	including a punctuation mark and quotation marks, as this sequence
	can occur both within and at the end of sentences.

\item[9.2\%] false negative resulting from an abbreviation followed by
	quotation marks;  related to the previous two types.

\item[9.8\%] false positive or false negative resulting from presence 
	of ellipsis (...), which
	can occur at the end of or within a sentence.

\item[9.9\%] miscellaneous errors, including extraneous characters
	(dashes, asterisks, etc.), ungrammatical sentences, misspellings,
	and parenthetical sentences.
\end{itemize}
\end{itemize}
\end{center}

The results presented above (409 errors) are obtained when both $t_0$
and $t_1$ are set at $0.5$.  Adjusting the sensitivity thresholds
decreases the number of punctuation marks which are mislabeled by the
method.  For example, when the upper threshold is set at $0.8$ and
the lower threshold at $0.2$, the network places 164 items between the
two.  Thus when the algorithm does not have enough evidence
to classify the items, some mislabeling can be avoided.\footnote{We
will report on results of varying the thresholds in future work.}

We also experimented with different context sizes and numbers of hidden
units, obtaining the results
shown in Tables \ref{contexts} and \ref{hidden}.
All results were found using the same training set of
573 items, cross-validation set of 258 items, and mixed-case test 
set of 27,294 items. 
The ``Training Error'' is one-half the sum of all the errors 
for all 573 items in the training set, where the ``error'' is the difference
between the desired output and the actual output of the neural net.
The ``Cross Error'' is the equivalent value for the cross-validation
set.  These two error figures give an indication
of how well the network learned the training data before stopping.

\begin{table*}
\centering
\begin{tabular}{| r r r r r r |} \hline
Context  & Training    & Training & Cross & Testing & Testing\\
Size     &    Epochs      & Error    & Error         & Errors   & Error (\%)\\ \hline \hline
4-context & 1731  & 1.52 & 2.36 & 1424 & 5.22\% \\
6-context & 218   & 0.75 & 2.01 & 409  & 1.50\% \\
8-context & 831   & 0.043 &1.88 & 877 & 3.21\% \\ \hline
\end{tabular}
\caption{\label{contexts} Results of comparing context sizes (2 hidden units).}
\end{table*}

\begin{table*}
\centering
\begin{tabular}{| r r r r r r |} \hline
\# Hidden  & Training    & Training & Cross & Testing & Testing\\
Units    &    Epochs  & Error    & Error    & Errors   & Error (\%)\\ \hline \hline

1 &  623 & 1.05 & 1.61 & 721 & 2.64\% \\
2 & 216  & 1.08 & 2.18 & 409 & 1.50\% \\
3 & 239  & 0.39 & 2.27 & 435 & 1.59\% \\
4 & 350  & 0.27 & 1.42 & 1343 & 4.92\% \\ \hline
\end{tabular}
\caption{\label{hidden} Results of comparing hidden layer sizes
(6-context).  Training was done  on 573 items, using a cross
validation set of 258 items.}
\end{table*}

We observed that a net with fewer hidden units results in a drastic
decrease in the number of false positives and a corresponding increase
in the number of false negatives.  Conversely, increasing the number
of hidden units results in a decrease of false negatives (to zero) and an
increase in false positives.  A network with 2 hidden units produces
the best overall error rate, with false negatives and false positives
nearly equal.

 From these data we concluded that a context of six surrounding tokens
and a hidden layer with two units worked best for our test set.

After converting the training, cross-validation and test texts to a
lower-case-only format and retraining, the network was able to
successfully disambiguate $96.2\%$ of the boundaries in a
lower-case-only test text.  Repeating the procedure with an
upper-case-only format produced a $97.4\%$ success rate.  Unlike most
existing methods which rely heavily on capitalization information, the
network method is reasonably successful at disambiguating single-case
texts.

\section{Discussion and Future Work}
\label{summary}

We have presented an automatic sentence boundary labeler which uses
probabilistic part-of-speech information and a simple neural network
to correctly disambiguate over $98.5\%$ of sentence-boundary
punctuation marks.  A novel aspect of the approach is its use of prior
part-of-speech probabilities, rather than word tokens, to represent
the context surrounding the punctuation mark to be disambiguated.
This leads to savings in parameter estimation and thus training time.
The stochastic nature of the input, combined with the inherent
robustness of the connectionist network, produces robust results.  The
algorithm is to be used in conjunction with a part-of-speech tagger,
and so assumes the availability of a lexicon containing prior
probabilities of parts-of-speech.  The network is rapidly trainable
and thus should be easily adaptable to new text genres, and is very
efficient when used in its labeling capacity.  Although the systems of
Wasson and Riley \shortcite{riley89} report slightly better error
rates, our approach has the advantage of flexibility for application
to new text genres, small training sets (and hence fast training
times), (relatively) small storage requirements, and little manual
effort.  Futhermore, additional experimentation may lower the error
rate.

Although our results were obtained using an English lexicon and text,
we designed the boundary labeler to be equally applicable to other
languages, assuming the accessibility of lexical part-of-speech
frequency data (which can be obtained by running a part-of-speech
tagger over a large corpus of text, if it is not available in the
tagger itself) and an abbreviation list.  The input to the neural
network is a language-independent set of descriptor arrays, so
training and labeling would not require recoding for a new language.
The heuristics described in Section \ref{algorithm} may need to be
adjusted for other languages in order to maximize the efficacy of
these descriptor arrays.  

Many variations remain to be tested. We plan to: (i) test the approach
on French and perhaps German, (ii) perform systematic studies on the
effects of asymmetric context sizes, different part-of-speech
categorizations, different thresholds, and larger descriptor arrays,
(iii) apply the approach to texts with unusual or very loosely
constrained markup formats, and perhaps even to other markup
recognition problems, and (iv) compare the use of the neural net with
more conventional tools such as decision trees and Hidden Markov
Models.

\medskip
\noindent
{\bf Acknowledgements} The authors would like to acknowledge valuable
advice, assistance, and encouragement provided by Manuel
F\"{a}hndrich, Haym Hirsh, Dan Jurafsky, Terry Regier, and Jeanette
Figueroa.  We would also like to thank Ken Church for making the PARTS
data available, and Ido Dagan, Christiane Hoffmann, Mark Liberman, Jan
Pedersen, Martin R\"{o}scheisen, Mark Wasson, and Joe Zhou for
assistance in finding references and determining the status of related
work.  Special thanks to Prof. Franz Guenthner for introducing us to
the problem.

The first author was sponsored by a GAANN fellowship; the second
author was sponsored in part by the Advanced Research Projects Agency
under Grant No. MDA972-92-J-1029 with the Corporation for National
Research Initiatives (CNRI) and in part by the Xerox Palo Alto
Research Center (PARC).

\end{document}